\documentclass{llncs}

\usepackage{amsfonts}
\usepackage{graphicx,graphics}
\usepackage{color}
\usepackage{url}
\usepackage{listings}
\usepackage{subfigure}
\usepackage{pgfplots,pgfplotstable,booktabs}

\newcommand{\ocp}{\textsc{Objective-CP}}
\newtheorem{defn}{Definition}
\newtheorem{exmp}{Example}

\begin{document}

\title{Domain Views for Constraint Programming}
\author{P. Van Hentenryck\inst{1} \and
 L. Michel\inst{2} 
}

\institute{
 NICTA, Australia, RI 02912 \\
 \and
 University of Connecticut, Storrs, CT 06269-2155
}
\maketitle

\begin{abstract}
  Views are a standard abstraction in constraint programming: They
  make it possible to implement a single version of each constraint,
  while avoiding to create new variables and constraints that would
  slow down propagation. Traditional constraint-programming systems
  provide the concept of {\em variable views} which implement a view
  of the type $y = f(x)$ by delegating all (domain and constraint)
  operations on variable $y$ to variable $x$. This paper proposes the
  alternative concept of {\em domain views} which only delegate domain
  operations. Domain views preserve the benefits of variable views but
  simplify the implementation of value-based propagation. Domain views
  also support non-injective views compositionally, expanding the
  scope of views significantly. Experimental results demonstrate the
  practical benefits of domain views. 
\end{abstract}

\section{Introduction}

Constraint programming systems provide rich libraries of constraints,
each of which models some specific structure useful across a wide
range of applications. These constraints are important both from a
modeling standpoint, as they make it possible to state problems at a
high level of abstraction, and from an efficiency standpoint, as they
allow dedicated algorithms to exploit the specific structure.  The
global constraint catalog \cite{GlobalConstraintCatalog} in fact lists
about 354 global constraints at the time of writing. In addition, each
of these constraints potentially come in many different forms as they
can be applied, not only on variables, but also on expressions
involving variables. 

This large number of variants presents a challenge for system
developers who must produce, validate, optimize, and maintain each
version of each constraint. To avoid the proliferation of such
variants, system developers often prefer to design a unique variant
over variables and introduce new variables and constraints to model
the more complex cases. For instance, a constraint
\[
{\it alldifferent}(x_1 + 1,\ldots,x_n + n)
\]
can be modeled by a system of constraints 
\[
\{{\it alldifferent}(y_1,\ldots,y_n), y_1 = x_1 + 1, \ldots ,y_n = x_n + n\}
\]
where the $y_i$'s are new variables. This approach keeps the system
core small but introduces an overhead in time and space. Indeed, the
new constraints must be propagated through the constraint engine and
the system must maintain additional domains and constraints,
increasing the cost of propagation and the space requirements. 

Over the years, system designers have sought ways to mitigate this
difficulty and proposed several solutions of varying complexity.
Prolog-style languages offered
indexicals~\cite{ccfd92,DBLP:conf/plilp/CarlssonOC97} while C++
libraries like Ilog Solver~\cite{SolverManual} introduces the concept
of variable views. For an injective function $f$ and a variable (or a
view) $x$, a variable view $y$ enforces the equivalent of the
constraint
$
y = f(x)
$
but it does not introduce a new variable and a new constraint:
Instead, it delegates all domain and constraint operations (the ability to wake constraints)
on $y$ to $x$, sometimes after applying $f^{-1}$. These variable views remove
the time and space overhead mentioned above and keep the solver kernel
small, thus giving us a valuable abstraction for constraint
programming. Recently, \cite{SchulteT08,Gecode13} demonstrated how variable
views can be implemented in terms of C++ templates, providing further
improvement in speed and memory usage. The idea is to use parametric
polymorphism to allow for code reuse and compile-time optimizations
based on code expansion and inlining. \cite{Gecode13}
demonstrates that variable views provide significant software
engineering benefits as well as great computational improvements over
the basic approach using new variables and constraints.

This paper aims at expanding the scope of constraint-programming views
with an extremely simple abstraction: The concept of {\em domain
  views} which only delegate domain operations. Domain views preserve
the benefits of variable views but simplify the implementation of
value-based propagation, i.e., the propagation of events of the form
$\langle c, x, v \rangle$, meaning that constraint $c$ must be
propagated because variable $x$ has lost value $v$ (e.g.,
\cite{AC5,Comet10}). The key benefit of domain views is to support
{\em non-injective views} elegantly and compositionalls. Domain views
can also be implemented using parametric polymorphism and hence are
fully compatible with the compilation techniques in \cite{Gecode13}.

The rest of the paper is organized as follows. Sections
\ref{sec:back}, \ref{section:views}, and \ref{section:variable-views}
present the preliminaries on constraint programming and on
views. Section presents the implementation of variable views. Section
\ref{section:domain-views} introduces the concept of domain
views. Section \ref{section:noninjective-views} demonstrates how to
generalize domain views to the case where the function $f$ is not
injective. Section \ref{section:monotone-views} briefly discusses how
to exploit monotonicity and anti-monotonicity. Section
\ref{section:empirical-results} presents experimental results. Section
\ref{section:related} discusses related work on advisors
\cite{Advisors07} and Section \ref{section:conclusion} concludes the
paper.

\section{Preliminaries}
\label{sec:back}

A constraint-programming system is organized around a 
queue of events ${\cal Q}$ and its main component is an engine propagating
constraints in the queue, i.e.,
\begin{lstlisting}
while $\neg$ empty(${\cal Q}$) do
    propagate(pop(${\cal Q}$));
\end{lstlisting}
For simplicity, we only consider two types of events: $\langle c, x
\rangle$ and $\langle c, x, v \rangle$. An event $\langle c, x
\rangle$ means that constraint $c$ must be propagated because the
domain of variable $x$ has been shrunk. An event $\langle
c, x, v \rangle$ means that constraint $c$ must be propagated because
the value $v$ has been removed from the domain of variable $x$.
Events of the form $\langle c, x \rangle$ are sometimes called {\em
  variable-based propagation}, while those of the form $\langle c, x,
v \rangle$ are sometimes called {\em value-based propagation}. Note
that some systems also implement what is called {\em constraint-based
  propagation}, where the event simply consists of constraint to
propagate without additional information.  We do not discuss
constraint-based propagation here since it is easier to handle.

The propagation of a constraint may change the domains of some
variables and thus introduce new events in the queue. As a result, a
variable $x$ not only maintains its domain $D(x)$ but also keeps track
of the constraints it appears in so that the proper events can be
inserted in the queue. As a result, a variable $x$ is best viewed as a
triple $\langle D, SC, SC_v \rangle$, where $D$ is the domain of the
variable, $SC$ is the set of constraints involving $x$ that use
variable-based propagation, and $SC_v$ is the set of constraints
involving $x$ that use value-based propagation. If $x$ is a variable,
we use $D(x)$, $SC(x)$ and $SC_v(x)$ to denote these three components.

\begin{figure}[t]
\begin{lstlisting}
interface Variable
  bool member(${\cal V}$ v);
  bool remove(${\cal V}$ v);
  void watch(${\cal C}$ c);
  void watchValue(${\cal C}$ c);
  void wake;
  void wakeValue(${\cal V}$ v);
\end{lstlisting}
\vspace{-4mm}
\caption{The Variable Interface.}
\label{fig:variable-interface}
\end{figure}

For simplicity, a variable in this paper implements the interface
depicted in Figure \ref{fig:variable-interface}, where ${\cal V}$
denotes the set of values considered (e.g., integers or reals) and
${\cal C}$ the set of constraints. For a variable $x$, method {\tt member(v)} tests
$v \in D(x)$, method {\tt remove(v)} implements 
$
D(x) := D(x) \setminus \{v\}
$
and returns true if the resulting domain is not empty, method {\tt
  watch(c)} registers constraint $c$ for variable-propagation, and
method {\tt watchValue(c)} registers constraint $c$ for
value-propagation. The {\tt wake} methods are used for creating new
events in the queue. Method {\tt wake} must implement 
$
{\cal Q} := {\cal Q} \cup \{ \langle c, x \rangle \ \mid \ c \in SC(x) \}
$
while method {\tt wakeValue(v)} must implement
$
{\cal Q} := {\cal Q} \cup \{ \langle c, x, v \rangle \ \mid \ c \in SC_v(x) \}.
$
With our conventions, a variable can be implemented as depicted in
Figure \ref{fig:var}.

\begin{figure}[t]
\begin{lstlisting}
implementation DomainVariable 
   $\{{\cal V}\}$ $D$;
   $\{{\cal C}\}$ $SC$;
   $\{{\cal C}\}$ $SC_v$;

DomainVariable($\{{\cal V}\}$ $D_o$) { $D$ := $D_o$; $SC$ := $\emptyset$; $SC_v$ := $\emptyset$; }
bool member(${\cal V}$ $v$) { return $v \in D$;}
bool remove(${\cal V}$ $v$) {
    if $v \in D$
       $D := D \setminus \{ v \}$;
       wake();
       wakeValue(v);
}
void watch(${\cal C}$ $c$)        { $SC := SC \cup \{c\}$; }
void watchValue(${\cal C}$ $c$) { $SC_v := SC_v \cup \{c\}$; }
void wake()          { ${\cal Q} := {\cal Q} \cup \{ \langle c, this \rangle \: | \: c \in SC \}$;}
void wakeValue(${\cal V}$ $v$)    { ${\cal Q} := {\cal Q} \cup \{ \langle c, this, v \rangle \: | \: c \in SC_v \}$;}
\end{lstlisting}
\vspace{-4mm}
\caption{The Implementation of a Domain Variable}
\label{fig:var}
\vspace{-4mm}
\end{figure}
\section{Views}
\label{section:views}

The purpose of this paper is to define and implement abstractions for
constraints of the form $y = \psi(x)$. In a first step, the paper
focuses on injective views, i.e., views in which function $\psi$ is
injective, which is the functionality provided by many
constraint-programming solvers.

\begin{defn}[Injective Function]
A function $\psi: D \rightarrow {\cal V}$ is injective if
\[
\forall v, v' \in D: \psi(v) = \psi(v') \Rightarrow v = v'.
\]
The inverse $\psi^{-1}: {\cal V} \rightarrow D_{\bot}$ of injective function $\psi$ is defined as
\[
\psi^{-1}(w) = \left\{\begin{array}{ll}
v & \mbox{ if } v \in D \ \wedge \ \psi(v) = w \\
\bot & \mbox{ otherwise} 
\end{array}\right.
\]
where $D_{\bot} = D \cup \{ \bot \}.$
\end{defn}

\noindent
Note that the definition of $\psi^{-1}$ is a specification: An actual
implementation uses a dedicated implementation of $\psi^{-1}$ as the
following two examples illustrate.

\begin{exmp}[Shift View]
  Consider the view $y=x+c$ where $c$ is an integer and $x$ and $y$
  are integer variables. Function $\psi: \mathbb{Z} \rightarrow \mathbb{Z}$ 
  can be specified (using lambda calculus notation~\cite{Barendregt84}) as $\lambda k.k + c$. Its inverse $\psi^{-1}:
  \mathbb{Z} \rightarrow \mathbb{Z}$ is defined as $\lambda k.k - c$.
\end{exmp}

\begin{exmp}[Affine View]
  Consider the view $y = a x + b$ where $a, b \in \mathbb{Z}$ and $x, y$
  are integer variables. $\psi: \mathbb{Z} \rightarrow \mathbb{Z}$ is $\lambda k. a k + b$. Its inverse $\psi^{-1}: \mathbb{Z}
  \rightarrow \mathbb{Z}$  is
\[
\psi^{-1} = \left\{\begin{array}{ll}
\lambda k. (k - b)/a & \mbox{ if } (k - b) \mbox{ mod } a = 0 \\
\lambda k. \bot & \mbox{ otherwise.}
\end{array}\right.
\]
\end{exmp}

\noindent
Views must be compositional and make it possible to state a view over
a view.

\section{Variable Views}
\label{section:variable-views}

The fundamental idea of variable views, implemented in many systems,
is to delegate all domain and constraint operations of variable $y$ to
variable $x$. A variable view thus implements an {\em adapter pattern}
that stores neither domain nor sets of constraints. The variable view
simply stores a reference to variable $x$ and delegates all domain and
constraint operations to $x$, possibly after applying function $\psi$
or $\psi^{-1}$ on the arguments. Informally speaking, the membership
test $w \in D(y)$ becomes $\psi^{-1}(w) \in D(x)$, the removal
operation proceeds similarly and variable $x$ also watches all the
constraints of $y$.

The only difficulty in variable views comes from the fact that
variable $x$ now needs to watch constraints on both $x$ and $y$. For
variable-based propagation, it is necessary to remember which variable
is being watched for each constraint and the set $SC$ now consists of
pairs $\langle c, z \rangle$ where $c$ is a constraint and $z$ is a
variable. For value-based propagation, it is necessary to store the
function $\psi$ since it must be applied when method {\tt wakeValue}
is applied. Hence the set $SC_v$ now contains triples of the form
$\langle c, z, \psi \rangle$. These generalizations are necessary,
since when a value $v$ is removed from the domain of $x$, the
value-based events for variable $y$ must be of the form $\langle c, y,
\psi(v) \rangle$.

The implementation of variables to support variable views is shown in
Figure \ref{fig:variable-implementation-for-vviews} where ${\cal X}$
denotes the set of variables/views and ${\cal F}$ the set of first-order
functions. Observe the types of $SC$ and $SC_v$ in lines 3--4, the new
methods in lines 14--15 allow to watch a constraint $c$ for a view
$y$, the matching redefinition of the {\tt watch} methods, and the 
{\tt wake} methods that store additional information
in the queue by applying the stored
function $\psi$ on value $v$ (line 20).

\begin{figure}[t]
\begin{lstlisting}
implementation DomainVariable 
   $\{{\cal V}\}$ $D$;
   $\{\langle {\cal C}, {\cal X}\rangle\}$ $SC$;
   $\{\langle {\cal C}, {\cal X}, {\cal F}\rangle\}$ $SC_v$;

DomainVariable($\{{\cal V}\}$ $D_o$) { $D$ := $D_o$; $SC$ := $\emptyset$; $SC_v$ := $\emptyset$;}
bool member(${\cal V}$ $v$) { return $v \in D$;}
bool remove(${\cal V}$ $v$) {
    if $v \in D$
       $D := D \setminus \{ v \}$;
       wake();
       wakeValue(v);
}
void watch(${\cal C} \ c,{\cal X} \ y$)                  { $SC  := SC \cup \{\langle c, y \rangle\}$; }
void watchValue(${\cal C} \ c,{\cal X} \ y,{\cal F} \ \psi$)        { $SC_v := SC_v \cup \{\langle c, y, \psi \rangle\}$;}
void watch(${\cal C}$ $c$)                 { watch($c$,this); }
void watchValue(${\cal C}$ $c$)            { watch($c$,this,$\lambda k.k$); }

void wake()        { ${\cal Q} := {\cal Q} \cup \{ \langle c,x\rangle \: | \: \langle c,x\rangle \in SC \}$;}
void wakeValue(${\cal V}$ $v$){ ${\cal Q} := {\cal Q} \cup \{ \langle c,x,\psi(v)\rangle \: | \: \langle c,x,\psi \rangle \in SC_v \}$;}
\end{lstlisting}
\vspace{-4mm}
\caption{The Domain Variable for Variable Views.}
\label{fig:variable-implementation-for-vviews}
\vspace{-4mm}
\end{figure}

Figure~\ref{fig:variable-views} depicts a template for variable views
in terms of an injective function $\psi$. A shift view specialization
is shown in Figure~\ref{fig:variable-view-shift}. Observe that
variable views do not store a domain nor constraint sets. Methods {\tt
  member} and {\tt remove} apply $\psi^{-1}$ as mentioned earlier with
only the addition of a test for the $\bot$ case. Methods {\tt watch}
and {\tt watchValue} (lines 10--11) state a view on the view
itself. In particular, line 11 illustrates the need for function
composition in the case of value propagation.

\begin{figure}[t]
\begin{lstlisting}
implementation VariableView<$\psi$> 
   ${\cal X}$  x;
VariableView(${\cal X}$ $\_x$) { $x$ := $\_x$; }
bool member(${\cal V}$ $v$) {
  if $\psi^{-1}(v) \neq \bot$ return $x$.member($\psi^{-1}(v)$);else return false; 
}
bool remove($\mathbb{Z}$ $v$) { 
  if $\psi^{-1}(v) \neq \bot$ return $x$.remove($\psi^{-1}(v)$);else return true;
}
void watch(${\cal C}$ $c$,${\cal X}$ $y$)               { $x$.watch($c,y$); }
void watchValue(${\cal C}$ $c$,${\cal X}$ $y$,${\cal F}$ $\phi$) { $x$.watchValue($c$,$y$,$\phi \circ \psi$);}
void watch(${\cal C}$ $c$)        { $x$.watch($c$,this); }
void watchValue(${\cal C}$ $c$) { $x$.watchValue($c$,this,$\psi$); }
\end{lstlisting}
\vspace{-4mm}
\caption{The Template for Variable Views.}
\label{fig:variable-views}
\vspace{-2mm}
\end{figure}

The instantiation for shift views in Figure
\ref{fig:variable-view-shift} highlights some interesting
points. First, there is no need for a $\bot$ test, since the inverse
of $\psi$ is always in the domain of $\psi$. Second, value-based
propagation requires the use of first-order functions (see lines 9 and
12) or objects implementing the same functionalities. In contrast,
methods {\tt member} and {\tt remove} ``inline'' function $\phi^{-1}$
in the code, which is never stored or passed as a parameter.

\begin{figure}[t]
\begin{lstlisting}
implementation VariableShiftView
   ${\cal X}$  $x$;
   $\mathbb{Z}$   $c$;
VariableShiftView(${\cal X}$ $\_x$,$\mathbb{Z}$ $\_c$) { $x$ := $\_x$; $c$ := $\_c$; }
bool member($\mathbb{Z}$ $v$) {  return $x$.member(v-c); }
bool remove($\mathbb{Z}$ $v$) {  return $x$.remove(v-c); }
void watch(${\cal C}$ $c$,${\cal X}$ $y$) {  $x$.watch($c,y$); }
void watchValue(${\cal C}$ $c$,${\cal X}$ $y$,$\mathbb{Z} \rightarrow \mathbb{Z}$ $\phi$) { $x$.watchValue($c$,$y$,$\phi \circ (\lambda k. k+c)$);}
void watch(${\cal C}$ $c$)        { $x$.watch($c$,this); }
void watchValue(${\cal C}$ $c$) { $x$.watchValue($c$,this,$\lambda k. k+c$); }
\end{lstlisting}
\vspace{-4mm}
\caption{A Variable View for Shift Views.}
\label{fig:variable-view-shift}
\vspace{-4mm}
\end{figure}

\paragraph{Optimization} 

Variable views now stores tuples $\langle c, z, \psi \rangle$ for
value-based propagation. Observe however that $z$ is an object so that
it is possible to use it to compute function $\psi$. This only
requires the view to provide a method {\tt map} that maps the value
$v$ through $\psi$.  Lines 15 and 20 in Figure
\ref{fig:variable-implementation-for-vviews} become
\begin{lstlisting}
void watchValue(${\cal C} \ c,{\cal X} \ y$)        { $SC_v := SC_v \cup \{\langle c, y \rangle\}$;}
void wakeValue(${\cal V}$ $v$)        { ${\cal Q} := {\cal Q} \cup \{ \langle c,x,x.\mbox{\tt map(}v\mbox{\tt )}\rangle \: | \: \langle c,x \rangle \in SC_v \}$;}
\end{lstlisting}
The \texttt{map} method on standard variables is defined as
\begin{lstlisting}
${\cal V}$ map(${\cal V}$ $v$) { return $v$;}
\end{lstlisting}
and its definition on views (defined over variable $x$ with injective function $\psi$) is 
\begin{lstlisting}
${\cal V}$ map(${\cal V}$ $v$) { return $\psi(\mbox{\tt x.map(}v{\mbox{\tt )}})$;}
\end{lstlisting}
Observe the recursive call, since views can be posted on views. This
optimization clutters a bit the API of variables and views but only
minimally. 

Variable views are an important concept in constraint programming for
injective functions. For constraint-based and variable-based
propagation, the implementation is simple and efficient, although it
requires to upgrade slightly the data structure to watch
constraints. For value-based propagation, the implementation is a bit
more cumbersome. It requires a generalization of the constraint queue
and the addition of a {\tt map} method on variables and views to avoid
manipulating first-order functions. Domain views provide an extremely
simple alternative, which also has the benefits of supporting
non-injective functions elegantly.

\section{Domain Views}
\label{section:domain-views}

The key idea behind domain views is to delegate only domain operations
from variable $y$ to variable $x$: The view for $y$ maintains its own constraints to
watch. This removes the need to manipulate first-order functions. To
implement domain views, traditional variables (and views) must store
which variables are viewing them. When their domains change, they
must notify their views. 

\begin{figure}[t]
\begin{small}
\begin{lstlisting}
implementation DomainVariable 
   $\{{\cal V}\}$ $D$;
   $\{{\cal C}\}$ $SC$;
   $\{{\cal C}\}$ $SC_v$;
   $\{{\cal X}\}$ ${\it Views}$;

DomainVariable($\{{\cal V}\}$ $D_o$) { $D$ := $D_o$;$SC$ := $\emptyset$;$SC_v$ := $\emptyset$;${\it Views}$ := $\emptyset$;}
void addView(${\cal X}$ $x$) { ${\it Views}$ := ${\it Views}$ $\cup$ $\{x\}$; }
bool member(${\cal V}$ $v$) { return $v \in D$;}
bool remove(${\cal V}$ $v$) {
    if $v \in D$
       $D := D \setminus \{ v \}$;
       wake();
       wakeValue(v);
       forall $y \in {\it Views}$
         y.wake();
         y.wakeValue(v);
}
void watch(${\cal C}$ $c$)        { $SC := SC \cup \{c\}$;}
void watchValue(${\cal C}$ $c$) { $SC_v := SC_v \cup \{c\}$;}
void wake()          { ${\cal Q} := {\cal Q} \cup \{ \langle c, this \rangle \: | \: c \in SC \}$;}
void wakeValue(${\cal V}$ $v$)    { ${\cal Q} := {\cal Q} \cup \{ \langle c, this, v \rangle \: | \: c \in SC_v \}$;}
\end{lstlisting}
\end{small}
\vspace{-4mm}
\caption{The Domain Variable for Domain Views.}
\label{fig:variable-implementation-for-dviews}
\vspace{-4mm}
\end{figure}

\begin{figure}[t]
\begin{lstlisting}
implementation DomainView<$\psi$> 
   ${\cal X}$  $x$;
   $\{{\cal C}\}$ $SC$;
   $\{{\cal C}\}$ $SC_v$;
   $\{{\cal X}\}$ ${\it Views}$;
DomainView(${\cal X}$ $\_x$)      { $SC$ := $\emptyset$; $SC_v$ := $\emptyset$; ${\it Views}$ := $\emptyset$;}
void addView(${\cal X}$ $x$) { ${\it Views}$ := ${\it Views}$ $\cup$ $\{x\}$; }
bool member(${\cal V}$ $v$) {
  if $\psi^{-1}(v) \neq \bot$ return $x$.member($\psi^{-1}(v)$);else return false; 
}
bool remove($\mathbb{Z}$ $v$) { 
  if $\psi^{-1}(v) \neq \bot$ return $x$.remove($\psi^{-1}(v)$);else return true;
}
void watch(${\cal C}$ $c$)        { $SC := SC \cup \{c\}$; }
void watchValue(${\cal C}$ $c$) { $SC_v := SC_v \cup \{c\}$; }
void wake() { 
  ${\cal Q} := {\cal Q} \cup \{ \langle c, this \rangle \: | \: c \in SC \}$;
  forall($y \in {\it Views}$) $y$.wake();
}
void wakeValue(${\cal V}$ $v$)  { 
  ${\cal Q} := {\cal Q} \cup \{ \langle c, this, v \rangle \: | \: c \in SC_v \}$;
  forall($y \in {\it Views}$) $y$.wakeValue($\psi(v)$);
}
\end{lstlisting}
\vspace{-4mm}
\caption{The Template for Domain Views.}
\label{fig:domain-views}
\vspace{-4mm}
\end{figure}

\begin{figure}[t]
\begin{lstlisting}
implementation DomainShiftView
   ${\cal X}$  $x$;
   $\{{\cal C}\}$ $SC$;
   $\{{\cal C}\}$ $SC_v$;
   $\{{\cal X}\}$ ${\it Views}$;
   $\mathbb{Z}$  $c$;
DomainShiftView(${\cal X}$ $\_x$,$\mathbb{Z}$ $\_c$) { 
  $SC$ := $\emptyset$; $SC_v$ := $\emptyset$; ${\it Views}$ := $\emptyset$;  $c$ := $\_c$; 
}
bool member($\mathbb{Z}$ $v$) {  return $x$.member(v-c); }
bool remove($\mathbb{Z}$ $v$) {  return $x$.remove(v-c); }
void watch(${\cal C}$ $c$)    { $SC := SC \cup \{c\}$; }
void watchValue(${\cal C}$ $c$) { $SC_v := SC_v \cup \{c\}$; }
void wake() { 
  ${\cal Q} := {\cal Q} \cup \{ \langle c, this \rangle \: | \: c \in SC \}$;
  forall($y \in {\it Views}$) $y$.wake();
}
void wakeValue(${\cal V}$ $v$)  { 
  ${\cal Q} := {\cal Q} \cup \{ \langle c, this, v \rangle \: | \: c \in SC_v \}$;
  forall($y \in {\it Views}$) $y$.wakeValue($v+c$);
}
\end{lstlisting}
\vspace{-4mm}
\caption{A Domain View for Shift Views.}
\label{fig:domain-view-shift}
\vspace{-4mm}
\end{figure}

Figure \ref{fig:variable-implementation-for-dviews} depicts the
revised implementation of domain variables to support domain views.
The variable now keeps its views (line 5) and provides a
method for adding a view (line 8). The only other change is in method
{\tt remove} in lines 16--18: The domain variable calls method
{\tt wake} and {\tt wakeValue} on its views to inform them of the loss
of value $v$ to let them schedule their own constraints. 

Figure~\ref{fig:domain-views} shows a template for domain views in
terms of an injective function $\psi$. A specialization for
 shift views is shown in Figure~\ref{fig:domain-view-shift}.
Observe first how the domain view maintains its own set of
constraints. It delegates its domain operations in methods {\tt
  member} and {\tt remove} in the same way as variable views, but it
does not delegate its {\tt watch} methods, which are similar to those
of a traditional domain variable. To implement views on
views, the {\tt wake} methods also wake the views (lines 16--20 and
21--25), using the function $\psi$ to send the appropriate value since
$v$ is the value removed from $D(x)$.  $D(x)$ may
be explicit (traditional variable) or implicit (views). The
shift view in Figure~\ref{fig:domain-view-shift} does not manipulate
first-order functions and inlines $\psi^{-1}$ in lines 9 and 12 and
$\psi$ in line 24.

Domain views provide an elegant alternative to variable views. They
remove the need to modify the data structure for watching constraint
and alleviate the need for the {\tt map} function, while preserving
the benefits of variable views and enabling more inlining for
value-based propagation. They are based on a simple idea: Only
delegating the domain operations. Instead of delegating constraint
watching, constraints are watched locally. It is interesting to
analyze the memory requirements of both approaches. Variable views
need to store variables in their constraint lists, which require space
proportional to the length of these lists. In contrast, domain views
only require a few pointers for their own lists, the constraints
themselves being present in both approaches albeit in different
lists. The viewed variables must also maintain the list of its views,
which is proportional to the number of views.

\section{Non-injective Views}
\label{section:noninjective-views}

We now generalize domain views to non-injective functions. 

\begin{defn}[Inverse of a Non-Injective Function]
The inverse $\psi^{-1}: {\cal V} \rightarrow 2^{D}_{\bot}$ of non-injective function $\psi: D \rightarrow {\cal V}$ is defined as
\[
\psi^{-1}(w) = \left\{\begin{array}{l}
\bot\mbox{ if } \not\exists \ v \in D: \psi(v) = w \\
\{ v \in D \ \mid \ \psi(v) = w \} \mbox{ otherwise.} 
\end{array}\right.
\]
\end{defn}

\begin{figure}[t]
\begin{lstlisting}
implementation NonInjectiveDomainView<$\psi$> 
   ${\cal X}$  $x$;
   $\{{\cal C}\}$ $SC$;
   $\{{\cal C}\}$ $SC_v$;
   $\{{\cal X}\}$ ${\it Views}$;
NonInjectiveDomainView(${\cal X}$ $\_x$) {
   $SC$ := $\emptyset$; $SC_v$ := $\emptyset$; ${\it Views}$ := $\emptyset$; 
}
void addView(${\cal X}$ $x$) { ${\it Views}$ := ${\it Views}$ $\cup$ $\{x\}$; }
bool member(${\cal V}$ $v$) {
  if $\psi^{-1}(v) \neq \bot$ return $\exists w \in \psi^{-1}(v)$ : $x$.member($w$); 
  else return false; 
}
bool remove(${\cal V}$ $v$) { 
  if $\psi^{-1}(v) \neq \bot$ 
     forall($w \in \psi^{-1}(v)$) if $\neg$ $x$.remove($w$) return false;
  return true;
}
void watch(${\cal C}$ $c$)        { $SC := SC \cup \{c\}$; }
void watchValue(${\cal C}$ $c$) { $SC_v := SC_v \cup \{c\}$; }
void wake() { 
  ${\cal Q} := {\cal Q} \cup \{ \langle c, this \rangle \: | \: c \in SC \}$;
  forall($y \in {\it Views}$) $y$.wake();
}
void wakeValue(${\cal V}$ $w$)  { 
  $v = \psi(w)$;
  if x.member(v)
     ${\cal Q} := {\cal Q} \cup \{ \langle c, this, v \rangle \: | \: c \in SC_v \}$;
     forall($y \in {\it Views}$) $y$.wakeValue($\psi(v)$);
}
\end{lstlisting}
\vspace{-4mm}
\caption{The Template for Non-Injective Domain Views.}
\label{fig:non-injective-domain-views}
\end{figure}

\noindent
Figure \ref{fig:non-injective-domain-views} gives the template for
non-injective views. There are only a few modifications compared to
the template for injective views.  The {\tt member} function must now
test membership for a set of values (line 12) and the {\tt remove}
function must remove a set of values (line 18). Finally, method {\tt
  wakeValue(w)} must test membership of $v=\psi(w)$, since there may
be multiple supports for $v$ in $D(x)$. 

The key advantage of domain views is that they own their
constraints. In the context of non-injective functions, this is
critical since only the view ``knows'' whether its constraints must be
scheduled for propagation. 

It is more difficult and less elegant, but not impossible, to
generalize variable views to support non-injective functions. Consider
what should happen for variable views.  For a view $y = f(x)$, when a
value $v$ is removed from the domain of $x$, it is no longer
sufficient to just use the {\tt map} function. The view must now
decide whether the value $f(v)$ is still supported for $y$. Moreover,
if we have a view $z = g(y)$ and variable $x$ is trying to decide
whether to schedule a constraint involving $z$, it must query $z$ to
find out whether the value $g(f(v))$ is still supported, which depends
on whether value $f(v)$ is still supported in variable $y$. Hence, to
implement non-injective functions in variable views, waking
constraints up must be conditional. It is necessary to implement a
method {\tt needToSchedule} on views to determine if the original
removal will actually remove a value on the views. Method {\tt
  wakeValue} now becomes
\begin{lstlisting}
void wakeValue(${\cal V}$ $v$) { 
   ${\cal Q} := {\cal Q} \cup \{ \langle c,x,x.\mbox{\tt map(}v\mbox{\tt )}\rangle \: | \: \langle c,x \rangle \in SC_v \; \& $ $x$.needToSchedule($v$)$\}$;}
\end{lstlisting}
The implementation of {\tt needToSchedule} must also be recursive
(like the {\tt map} function) to handle the case of views on views.
For space reasons, we let readers figure out the details on how to do
so correctly and {\em only note the conceptual simplicity of domain
  views.}\footnote{Method {\tt needToSchedule} must also update any
  internal state of the views. From a semantic standpoint, it would
  desirable to have another recursive method to notify the view that
  value $v$ has been removed and to update the state.}

\paragraph{Literal Views}
\label{sec:lit}

Reified constraints are a fundamental abstraction in constraint
programming.  For instance, In a magic series $s$ of length $n$, every
$s_i$ must satisfy $s_i = \sum_{j=0}^{n-1} (s_j=i)$, i.e., it states
that $s_i$ should be the number of occurrences of value $i$ in $s$
itself. To implement this behavior, one could rely on auxiliary
boolean variables $b_{ij} \Leftrightarrow s_j = i$. for every $i$ and
$j$ in $0..n-1$ leading to a quadratic number of boolean variables and
reified equality constraints. The reification $b \Leftrightarrow x =
i$ can be seen as a non-injective view and Figure
$\ref{fig:domain-view-reifed}$ describes its implementation. The view
uses two methods not described before: Method {\tt isBoundTo($i$)} on
variable $x$ holds if $D(x) = \{i\}$, while method {\tt bind($i$)}
succeeds if $i \in D(x)$ and reduces the domain $D(x)$ to
$\{i\}$. With these two functions, the implementation is direct with
the methods {\tt member}, {\tt remove}, and {\tt wakeValue} carried
out by case analysis on the value of the ``reified variable''.

\begin{figure}[t]
\begin{lstlisting}
implementation ReifedDomainView
   ${\cal X}$  $x$;
   $\{{\cal C}\}$ $SC$;
   $\{{\cal C}\}$ $SC_v$;
   $\{{\cal X}\}$ ${\it Views}$;
   $\mathbb{Z}$  $i$;
DomainReifiedView(${\cal X}$ $\_x$,$\mathbb{Z}$ $\_i$) { 
  $SC$ := $\emptyset$; $SC_v$ := $\emptyset$; ${\it Views}$ := $\emptyset$;  $i$ := $\_i$; 
}
bool member($\mathbb{Z}$ $v$) {   
  if $v = 0$ return $\neg x$.isBoundTo($i$); 
  else return $x$.member($i$); 
}
bool remove($\mathbb{Z}$ $v$) {   
   if $v = 0$ 
      return $x$.bind($i$);
   else return $x$.remove($i$);
}
$\ldots$
void wakeValue(${\cal V}$ $v$)  { 
   if $v = i$
      ${\cal Q} := {\cal Q} \cup \{ \langle c, this, 1 \rangle \: | \: c \in SC_v \}$;
      forall($y \in {\it Views}$) $y$.wakeValue($1$);
   else      
      if $\neg$member(0) 
         ${\cal Q} := {\cal Q} \cup \{ \langle c, this, 0 \rangle \: | \: c \in SC_v \}$;
         forall($y \in {\it Views}$) $y$.wakeValue($0$);
}
\end{lstlisting}
\vspace{-4mm}
\caption{A Domain View for Reified Views.}
\label{fig:domain-view-reifed}
\end{figure}

\paragraph{Modulo Views}

We now show a view for a constraint $y = x \mbox{ mod } k$ with $k \in \mathbb{Z}$. The view implementation maintains
the supports for each value $v \in D(y)$, i.e., 
\[
\forall v \in D(y) \: s_v = \{ w \:|\: w \in D(x) \wedge w \mbox{ mod } k = v\}
\] 
Figure \ref{fig:domain-view-modulo} depicts a sketch of a simple implementation. 

\begin{figure}[t]
\begin{lstlisting}
implementation ModuloDomainView
   $\ldots$
   int      k;
   $\{\mathbb{Z}\}$[]  S;   
DomainReifiedView(${\cal X}$ $\_x$,$\mathbb{Z}$ $\_k$) { $\ldots$ }
bool member($\mathbb{Z}$ $v$) {  return $S_v \neq \emptyset$; } 
bool remove($\mathbb{Z}$ $v$) {   
  forall($w \in S_v$)
     if $\neg$ $x$.remove($w$) return false;
  return true;
}
$\ldots$
void wakeValue(${\cal V}$ $w$)  { 
   $v$ := $w$ mod $k$;
   if $\neg$ member($v$)
       ${\cal Q} := {\cal Q} \cup \{ \langle c, this, v \rangle \: | \: c \in SC_v \}$;
       forall($y \in {\it Views}$) 
          $y$.wakeValue($v$);
}
\end{lstlisting}
\vspace{-4mm}
\caption{A Domain View for a Modulo Function.}
\label{fig:domain-view-modulo}
\end{figure}

%%%%%%%%%%%%%%%%%%%%%%%%%%%%%%%%%%%%%%%%%%%%%%%%%%%%%%%%%%%%%%%%%%%%%%%%%%%%%%%%
% Integer Views
%%%%%%%%%%%%%%%%%%%%%%%%%%%%%%%%%%%%%%%%%%%%%%%%%%%%%%%%%%%%%%%%%%%%%%%%%%%%%%%%

\section{Monotone and Anti-Monotone Views}
\label{section:monotone-views}

We briefly mention how to exploit monotone and anti-monotone properties to
perform additional operations such as {\tt updateMin} and {\tt
  updateMax}. These techniques are well-known and are only reviewed
here for completeness.

\begin{defn}[Monotone/AntiMonotone Function]
An injective function  $\psi$ is monotone if $\forall v,w : v \leq w \rightarrow \psi(v) \leq \psi(w)$. 
It is anti-monotone if $\forall v, w : v \leq w \rightarrow \psi(v) \geq \psi(w)$. 
\end{defn}

\noindent
If $\psi: \mathbb{Z} \rightarrow \mathbb{Z}$ is a monotone function and
$y$ is a view on $x$, then the update operations on bounds becomes 
\begin{lstlisting}
bool updateMin($\mathbb{Z}$ $v$) { return $x$.updateMin($\psi^{-1}(v)$); }
bool updateMax($\mathbb{Z}$ $v$) { return $x$.updateMax($\psi^{-1}(v)$); }
\end{lstlisting}
ignoring the case where $\psi^{-1}(v)$ is not well-defined. When $\psi$
is anti-monotone, they become
\begin{lstlisting}
bool updateMin($\mathbb{Z}$ $v$) { return $x$.updateMax($\psi^{-1}(v)$); }
bool updateMax($\mathbb{Z}$ $v$) { return $x$.updateMin($\psi^{-1}(v)$); }
\end{lstlisting}

\section{Empirical Evaluation}
\label{section:empirical-results}

We now describe experimental results to demonstrate the efficiency of
domain views. The experiments were run on MacOS X 10.8.3 running on a
Core i7 at 2.6Ghz, using the \ocp{} optimization system
\cite{ObjectiveCP13}. The complete implementation of the integer and
boolean variables, along with their domain and their views (including
literal views) is around 3,200 lines of code, which is similar to the
type of code reuse advertised for Gecode~\cite{Gecode13}. \ocp{}
pushes the methodology advocated in \cite{Gecode13} to the limit, only
supporting core constraints and using views to obtain more complex
versions. For instance, the CP solver in \ocp{} provides $\sum_{i=0}^n
x_i \leq b$ but not $\sum_{i=0}^n a_i \cdot x_i \leq b$.  Note that cost-based 
propagation for COP would, of course,  
mandate global constraints retaining the $a_i$. 
\ocp{}
supports value-based propagation and non-injective views, which
demonstrates the additional functionalities provided by domain
views. Note that the experiments only aim at demonstrating the
practicability of domain views: See \cite{Gecode13} for the benefits
of views. 

\paragraph{Benchmarks}

The implementation was validated with on a variety of benchmarks
relying on views. The experiments compare implementations with no
views, with the optimized variable views (with subtype polymorphism),
and domain views. When no-views are used, the implementation uses the
constraints and auxiliary variables introduced during the flattening
of the model.  The implementation uses the same models throughout and
the search space and pruning are always identical. For \texttt{bibd},
we follow \cite{Gecode13} and rewrite the boolean relations $a \wedge
b$ as $\neg \left( \neg a \vee \neg b\right)$ to ensure that the
system uses negation views. Specifically, \texttt{knapsack} use linear
equations $\sum_{i \in S} x_{i} = b$ and introduce views for the
coefficients. The Steel Mill \texttt{Slab} problem relies on literal
views for the color constraint on slab $s$~:~ $ \sum_{c \in Colors}
\vee_{o \in Orders[c]} (x_{o}=s) \leq 2.  $ \texttt{Debruijn} uses
both linear equations as well as reifications. \texttt{Langford} uses
affine views to ``shift'' indices within element
constraints. \texttt{Magicseries} clearly relies on
reifications. \texttt{Sport} is the classic sport scheduling benchmark
and uses global constraints.

\paragraph{Measurements}

The benchmarks use a simple first-fail heuristic as decomposition may
change the behavior of more advanced heuristics (e.g., WDEG) and these
experiments are only interested in assessing view implementations, not
inherent speed. Table~\ref{tab:benchmarks} offers a comparative view
of the results. It is based on 50 execution of each benchmark to
account for the inherent variability related to modern processor
technology. Columns $\mu(T_{cpu})$ and $\mu(T_{wc})$ give the average user-time or
wall-clock times in milliseconds. Columns $\sigma(T_{cpu})$ and
$\sigma(T_{wc})$ report the standard deviations for those run times.
Column $|M|$ reports the peak memory consumption in kilobytes for the
entire process. The measurement was taken at the level of the
\texttt{malloc} C-runtime function and includes all memory allocations
done by the executable. Finally, column $P.$ reports the number
of propagation events recorded by the engine (in thousands).

\begin{table}[t]
\begin{center}
 \pgfplotstableset{fixed zerofill,precision=1}
\noindent\begin{filecontents*}{summary.csv}
row,bench,type,mcpu,mw,sdcpu,sdwc,mpeak,sdpeak,mp
2,bibd(6),nv,1088.42,1130.46,222.114246205296,227.877536928796,45723967.68,76870.3166865296,1983638
1,bibd(6),dv,729.8,759.22,107.622165989376,110.990584266947,29898117.76,37026.4202211097,804196
3,bibd(6),vv,644.74,671.16,59.3845338210452,60.1848716498607,28756204.16,65143.6773931316,804921
8,knapsack(4),nv,8857.48,8873.9,180.870597558654,184.752481963942,1010672.96,731.759996340944,33207501
7,knapsack(4),dv,6767.96,6784.46,166.759017135261,176.089403313157,831468.48,1055.60556641045,2949428
9,knapsack(4),vv,6151.18,6164.52,109.154391686683,110.951771099898,807579.84,1554.72198919082,3061780
14,ais(30),nv,1341.76,1348.92,52.1600989106362,57.6080493241941,1368320.32,894.035689375738,2734177
13,ais(30),dv,1355.3,1361.54,31.8044086277872,32.444118362602,1368133.12,953.285032219148,2734177
15,ais(30),vv,1354.94,1362.36,26.0685203857457,26.5258903176285,1369471.04,1904.62801510383,2734177
17,sport,nv,4851.74,4864.24,111.331424685257,115.9999225897,2078548.8,3107.8473946723,3360730
16,sport,dv,4850.9,4864.3,179.366828548853,189.280404238187,2077998.72,3610.65574371081,3360730
18,sport,vv,4936.08,4949.52,231.776444014486,236.669944980648,2077684.8,4082.14653418157,3360730
20,langford(9/3),nv,6060.54,6076.06,298.776703870148,306.185978564316,1408477.76,1886.57991560368,54027018
19,langford(9/3),dv,7008.32,7026.58,316.768790979755,323.521698480573,1401280.96,2887.40890101015,56892930
21,langford(9/3),vv,6859.54,6877.92,242.690716925128,248.498662584173,1399169.92,2003.88622268566,56887091
23,debruijn(2/12),nv,7665.54,8437.34,153.385230643605,169.213258755854,639626867.52,121363.395947913,2557792
22,debruijn(2/12),dv,7292.22,8014.22,111.183850317519,144.067343946631,565776103.04,632.008555870787,946417
24,debruijn(2/12),vv,6935.24,7628.24,384.906220224286,422.395769536638,564011955.52,319.171007838898,966820
26,slab,nv,4845.2,4919.38,108.557517905844,115.385934487489,86110373.76,2929.09404786266,4402682
25,slab,dv,2243.46,2293.98,128.731520964354,138.757826034117,52966527.68,451.270742753512,908939
27,slab,vv,4509.76,4578.66,94.7069723528574,99.4085796814627,75703074.24,1234.43812815584,2967612
29,magicserie(300),nv,17951.18,18164.66,284.036551973761,300.770005696264,237181103.68,220.177056766891,30770809
28,magicserie(300),dv,8318.28,8443.62,191.931220673938,201.404952009323,124955868.16,155.816658353361,257238
30,magicserie(300),vv,14879.9,15087.96,429.060637453781,441.06535958809,234790278.72,38888.9461903707,20745063
\end{filecontents*}
\pgfplotstabletypeset[
 string replace*={dv}{\textsc{Domain-View}},
 string replace*={nv}{\textsc{No-View}},
 string replace*={vv}{\textsc{Var-View}},
 columns={bench,type,mcpu,mw,sdcpu,sdwc,mpeak,mp}, 
  columns/bench/.style ={string type,column name=$Bench$,column type=l},
  columns/type/.style={string type,column type=l},
  columns/mcpu/.style={column name=$\mu(T_{cpu})$,column type=r,precision=1},
  columns/sdcpu/.style={column name=$\sigma(T_{cpu})$,column type=r,precision=1},
  columns/mw/.style={column name=$\mu(T_{wc})$,column type=r,precision=1},
  columns/sdwc/.style={column name=$\sigma(T_{wc})$,column type=r,precision=1},
  columns/mpeak/.style={column name=$|M| (KB)$,preproc/expr=##1 / 1024,column type=r,precision=0,fixed},
  columns/mp/.style={column name=$P.(\times1000)$,preproc/expr=##1 / 1000,column type=r,precision=0},
  every head row/.style={before row=\toprule,after row=\midrule},
  every last row/.style={after row=\bottomrule},
  col sep=comma
  ]{summary.csv}
\end{center}
\vspace{-2mm}
\caption{Experimental Results on Variable and Domain Views.}
\label{tab:benchmarks}
\vspace{-8mm}
\end{table}

Without surprise, the results indicate that a minimalist kernel
\textit{must} use views to be competitive. The differences in memory
consumptions and running times are often quite significant when
contrasted with view-based implementations. For all benchmarks
involving only injective views, variable and domain views are
essentially similar in time and space efficiency. Given the standard
deviations, the differences in efficiency are not statistically
significant, although variable views are often slightly more
efficient. This is not always the case, as the sport-scheduling
problem indicates. The main benefit of domain views is to support
non-injective views simply and efficiently. This is particularly clear
on the benchmarks relying on reifications, i.e, \texttt{slab} and
\texttt{magicserie}.  The benefits are in terms of runtime and memory
consumption. The runtime benefits are quite substantial, as the
running time is halved on the Steel Mill Slab problem. The dramatic 
drop in the number of propagations is easily explained by the absence of 
constraints of the form $b \Leftrightarrow (x=v)$, yet, the same work is still
carried out by the view, albeit at a much lower overhead. 

In summary, the experimental results show that domain views do not add
any measurable overhead on injective views and bring significant
benefits on non-injective views, which they support elegantly.

% \begin{table}[t]
% {\scriptsize
% \begin{center}
% \noindent\pgfplotstabletypeset[
% 	columns={b,xc,xp,xm,xt}, 
% 	columns/b/.style ={string type,column name=$Bench$,column type=l},
% 	columns/xc/.style={column name=G($|C|$),
% 		preproc/expr=##1 * 100,
% 		postproc cell content/.append style={
%             /pgfplots/table/@cell content/.add={}{\%},
%         },column type=r},
% 	columns/xp/.style={column name=G(P),
% 		preproc/expr=##1 * 100,
% 		postproc cell content/.append style={
%             /pgfplots/table/@cell content/.add={}{\%},
%         },column type=r},
% 	columns/xm/.style={column name=G(M),
% 		preproc/expr=##1 * 100,
% 		postproc cell content/.append style={
%             /pgfplots/table/@cell content/.add={}{\%},
%         },column type=r},
% 	columns/xt/.style={column name=G(T),
% 		preproc/expr=##1 * 100,
% 		postproc cell content/.append style={
%             /pgfplots/table/@cell content/.add={}{\%},
%         },column type=r},
% 	every head row/.style={before row=\toprule,after row=\midrule},
% 	every last row/.style={after row=\bottomrule}
% ]{bench.csv}
% \end{center}
% }
% \vspace{-3mm}
% \caption{Comparison of Views and Decomposition.}
% \label{tab:comp}
% \end{table}

\section{Related Work}
\label{section:related}

It is important to contrast the variable and domain view
implementations proposed here with another approach using delta-sets
and advisors \cite{Advisors07,Gecode13}. Advisors are another way of
``simulate'' value-based propagation.\footnote{It is only a simulation
  since an advisor updates the constraint state but does not propagate
  a constraint itself. They are second-class citizens by choice in
  Gecode \cite{Advisors07}.} An advisor is associated with a variable
and a constraint and it modifies the state of the constraint directly
upon a domain modification for its variables. Advisors do not go
through the propagation queue but modify the state of their constraint
directly. This has both an advantage (speed) and an inconvenience,
since an advisor may be called while its constraint is propagating;
Hence some care must be exercised to maintain a consistent
state. Advisors also receive the domain change (called a delta set)
which they may query.

Advisors can be associated with variable views. The view must now be
upgraded to query, not only the domain, but also the delta sets. In
other words, the queries on the delta must transform the domain delta,
say $\{v_1,\ldots,v_n\}$, through the view to obtain
$\{\phi(v_1),\ldots,\phi(v_n)\}$. Gecode \cite{Gecode13} does not
compute delta sets exactly but approximates them by intervals
instead. A complete implementation of value-based propagation would
require the creation of these delta sets. Advisors and delta sets can
be used in the case of non-injective functions but that solution would
still go through the propagation queue and use a constraint. Indeed,
by design, advisors do not propagate constraints.

The key advantage of domain views in this context is their ability to
implement non-injective views without going through the propogation
queue.

\section{Conclusion}
\label{section:conclusion}

This paper reconsidered the concept of views, an important abstraction
provided by constraint-programming systems to avoid the proliferation
of constraints, while preserving the efficiency of dedicated
implementation. It proposed an alternative to the concept of variable
views, typically featured in constraint-programming systems. Contrary
to variable views, domain views only delegate domain operations and
maintains their own set of constraints to watch. Domain views simplify
the implementation of constraint-programming systems featuring
value-based propagation as they avoid manipulating first-order
functions (or objects implementing a similar functionality). They also
make it possible to implement, in simple ways, views featuring
non-injective functions. These are particularly useful for reified
constraints, which are also an important features of
constraint-programming systems. Experimental results demonstrate that
domain views introduce a negligible overhead (if any) over variable
views and that views over non-injective functions, which are elegantly
supported by domain views, provide significant benefits.

\newpage

\bibliographystyle{abbrv}

%\bibliography{local,../../biblio/biblio}

\end{document}